\pdfoutput=1

\documentclass[11pt]{article}
\usepackage{kotex}

\usepackage[]{ACL2023}

\usepackage{times}
\usepackage{latexsym}
\usepackage{adjustbox}
\usepackage{graphicx}

\usepackage[T1]{fontenc}

\usepackage[utf8]{inputenc}

\usepackage{microtype}

\usepackage{inconsolata}
\usepackage{setspace}
\usepackage{tabularx,ragged2e}
\newcolumntype{C}{>{\Centering\arraybackslash}X} 

\title{Structural Quality Metrics to Evaluate Knowledge Graph Quality}

\author{Sumin Seo, Heeseon Cheon, Hyunho Kim, Dongseok Hyun\\
  \texttt{sumin.seo@navercorp.com}, \texttt{heeseon.cheon@navercorp.com},\\ \texttt{kim.hh@navercorp.com}, \texttt{dustin.hyun@navercorp.com}}

\begin{document}
\maketitle
\begin{abstract}
This work presents six structural quality metrics that can measure the quality of knowledge graphs and analyzes four cross-domain knowledge graphs on the web (Wikidata, DBpedia, YAGO, Freebase) and Google Knowledge Graph, as well as `Raftel',  Naver's integrated knowledge graph. The `Good Knowledge Graph' should define detailed classes and properties in its ontology so that it can abundantly express knowledge in the real world. Also, instances and RDF triples should use the classes and properties actively. Therefore, we tried to examine the internal quality of knowledge graphs numerically by focusing on the structure of the ontology, which is the schema of knowledge graphs, and the degree of use thereof. As a result of the analysis, it was possible to find the characteristics of a knowledge graph that could not be known only by scale-related indicators such as the number of classes and properties.\\
\end{abstract}

\section{Introduction}

A knowledge graph is a data system that consists of RDF triples which are in the form of a \textit{`subject-predicate-object'} between entities in the real world and their relationships. For example, the fact that `the capital of Korea is Seoul' can be expressed as \textit{`Korea - capital - Seoul'}.

Ontology is a schema that defines the structure of a knowledge graph. An ontology defines the hierarchical relationship between classes and the properties that classes can have. Class is an abstract concept that encompasses entities with similar characteristics. For example, \textit{Seoul} is the instance(=entity) of the \textit{`City'} class, and \textit{Korea} is the instance of the \textit{`Country'} class. The hierarchical relationship between classes is expressed with \textit{`subclass of'} predicate between superclass, which means a higher concept, and subclass, which means a lower concept. In addition, \textit{`is-a'} relationship must be established between the two. For example, the \textit{`Book'} class and the \textit{`Movie'} class are subclasses of the \textit{`Creative Work'} class and are defined in the ontology with RDF triples, \textit{`Book - subclass of - Creative Work'} and  \textit{`Movie - subclass of - Creative Work'}.  Property is an attribute that each class can have, and in the case of the \textit{`Country'} class, it can have \textit{`capital'}, \textit{`population'}, and \textit{`president'} as properties. An ontology defines the properties that each class can have.

Knowledge graphs has an important role in search systems such as Google's knowledge panel (\citet{Zou_2020}), and recently attention is being drawn from NLP tasks such as Question Answering (\citet{10.1145/3289600.3290956}) or recommendation systems (\citet{9216015}), and explainable AI (\citet{TIDDI2022103627}). In this regard, Wikidata (\citet{42240}, Freebase(\citet{10.1145/1376616.1376746}, DBpedia (\citet{inproceedings}), and YAGO(\citet{10.1145/1242572.1242667}) are examined for various tasks in many studies.

This study outlines what makes a "good knowledge graph" and offers a metric to quantify it. In contrast to the current knowledge graph evaluation studies, which mainly focused on the size and distribution of the data, this study devised a measure to compare the quality between knowledge graphs from the viewpoint that structure (=ontology) is a key factor in determining the quality of knowledge graphs. Based on this indicator, we compared knowledge graphs on the web (Wikidata, Freebase, DBpedia, YAGO, Google Knowledge Graph) and Naver's knowledge graph Raftel.

\section{Related Works}

Ontology and knowledge graph evaluation methods can be categorized as follows.(\citet{brank2005survey}, \citet{Raad2015ASO}, \citet{Frber2018WhichKG}) 

\begin{itemize}
  \item \textit{gold standard evaluation}: a method of comparing knowledge graphs to high-quality knowledge graphs with the same topic.
  \item \textit{data driven evaluation}: a method that selects important words by extracting keywords from documents dealing with the same domain of knowledge graphs and measures how much information knowledge graphs contain.
  \item \textit{application/task based evaluation}: a method that evaluates the downstream task performance of the knowledge graph.
  \item \textit{user based evaluation}: a method that evaluates quality from the perspective of knowledge graph users.
  \item \textit{structure based evaluation}: a method that evaluates knowledge graphs through metrics that can reflect the structure or statistical properties of the ontology and knowledge graphs. (\citet{Lourdusamy2018ARO}, \citet{Tartir2005OntoQAMO})
  \item \textit{data quality evaluation}: a method that defines data quality with various point of views including \textit{accuracy} and \textit{consistency} of data and suggests numerical indicators to measure data quality.(\citet{10.3233/SW-170275}, \citet{Frber2018WhichKG})
\end{itemize}

Comparative studies on cross-domain knowledge graphs on the web are mainly focuses on structure-based evaluation and data quality evaluation. (\citet{Ringler2017OneKG}, \citet{heist2020knowledge}) First of all, the structure-based evaluation uses \textit{schema metric}, \textit{instance/knowledge base metric}, \textit{class metric}, \textit{graph metric} and \textit{complexity metric} to compare knowledge graphs. \textit{Schema metric} calculates the number of classes, the number of properties, and the number of properties per class, focusing on an ontology. \textit {Instance/knowledge base metric} calculates, for example, the average number of instances per class considering instances with an ontology. In the case of \textit{class metric}, the number of instances of the \textit{`Person'} class and the degree of connection with other classes is calculated to represent the characteristics of each class. \textit {Graph metric} is the application of basic statistics of graph theory such as cohesion and cardinality to knowledge graphs.

\citet{10.1145/3306446.3340822} analyzed the dimensions on which researches on the data quality of knowledge graphs concentrated according to the framework presented by \citet{10.1080/07421222.1996.11518099}. There are many works that evaluate knowledge graphs on data quality perspective: An \textit{accuracy} perspective which means how accurately the knowledge graph reflects real-world information (\citet{2f14de77803f4c3fbab040b6b0a63514}, \citet{Prasojo2016ManagingAC} ), a \textit{consistency} perspective that focuses on how consistent the data in the knowledge graph is(\citet{conf/icwsm/SpitzDRGG16}, \citet{10.1145/3274410}), an \textit{ease of understanding} perspective including how many languages a knowledge graph  provides(\citet{10.1145/3125433.3125465}), a \textit{interlinking} perspective which means how much a knowledge gaph can be connected to other knowledge graphs(\citet{10.1145/1242572.1242667}, \citet{Ringler2017OneKG}).

As a complement to the shortcomings of current structure-based metrics and data-quality-based assessment techniques, we introduce the \textit{structural quality metrics} as a measure of knowledge graph quality. Most of the researches comparing knowledge graphs on the web with structural metrics mainly focus on the size of knowledge graphs (number of RDF triple, class, and instance). In addition, numerical indicators related to structure simply describe the distribution of data such as the number of instances per class. It is easy to grasp the size and approximate structure of the knowledge graph with this approach, but difficult to evaluate either side as better in terms of quality. Graph metrics such as cohesion and cardinality can be used to evaluate the quality of knowledge graphs, such as how concisely organized knowledge graphs are and how rich relations a knowledge graph has. However, these metrics are not specialized for the quality of knowledge graphs. Studies based on data quality dimension analyzed knowledge graphs on the web in various aspects, but there are no quality indicators based on the knowledge graph structure, ontology. Therefore, we proposes \textit{structural quality metric} that can numerically represents the internal quality of knowledge graphs, focusing on the ontological structure and its degree of utilization.

\section{Data Introduction and Basic Statistics}
\subsection{Data Introduction and Data Preparation}

\begin{table*}[!ht]
\setlength\extrarowheight{2pt} 
\begin{tabularx}{\textwidth}{|C|C|C|C|}
\hline
                 & \textbf{subject} & \textbf{predicate} & \textbf{object}  \\\hline
\textbf{class} & <http://rdf.freebase.com /ns/base. birdinfo.parasitism> & <http://www.w3.org /1999/02/22 -rdf-syntax-n\#type> & <http://www.w3.org /2000/01/ rdf-schema\#Class>  \\
\hline
\textbf{predicate} & <http://rdf.freebase.com /ns/film. film.directed\_by>& <http://www.w3.org /2000/01/ rdf-schema\#domain>  & <http://rdf.freebase.com /ns/film.film>\\
\hline
\end{tabularx}
\caption{\label{freebase-preperation}Example of Freebase Ontology data extraction} 
\end{table*}

Wikidata, DBpedia, Freebase, YAGO, and Google Knowledge Graph, which are knowledge graphs analyzed in the work, have the following characteristics.(\citet{heeseon})

\begin{description}
\item [Wikidata] Wikidata was launched in 2012 by Wikimedia Deutschland, which provides information associated with Wikipedia and allows users to participate directly in creating and editing data. Although ontology information is not provided as a separate file, the hierarchical structure between entities can be known through \textit{subclass of} property for each entity.

\item [DBpedia] DBpedia is a knowledge graph launched in 2007 by Free University of Berlin and the University of Leipzig. It is created by automatically extracting structured information contained in Wikipedia, and it builds and manages its own ontology structure.

\item [Freebase] Freebase was launched by MetaWeb Technologies, Inc. in 2007 and merged into Wikidata by the Wikimedia Foundation and Google in 2015.(\citet{44818}) Ontology is constructed in a human readable manner with the structure of \textit{`domain/class/predicate'}.

\item [YAGO] Developed by Max Planck in 2007, data is generated by extracting information from Wikipedia infobox and WordNet in various languages. The ontology structure is built on WordNet.

\item [Google Knowledge Graph] Google KG was developed by Google in 2012 to understand the meaning of search terms in the search engine. Through the API, information such as name, description, image, and type which informs the class can be obtained for a specific keyword. There is no self-defined ontology, and the type is configured based on \textit{schema.org}. In 2014, Google introduced Knowledge Vault \cite{45634}, which integrates Wikipedia, YAGO, Microsoft's Satori, and Google Knowledge Base. Since the data is not published to the public, we used Google Knowledge Graph data accessible by API (Google Knowledge Graph Search API).
\end{description}

\begin{table*}[ht!]
\renewcommand{\arraystretch}{1.4}
\centering
\begin{adjustbox}{width=1\textwidth}
\small
\begin{tabularx}{\textwidth}{lllllll}
\hline
                      & Raftel & Wikidata    & DBpedia    & YAGO                                                                  & Google KG   & Freebase    \\
                      \hline
number of classes     & 273         & 59,662     & 804                                                                   & 266         & 910      & 53,091     \\
\hline
number of properties  & 607         & 7,476      & 21,607                                                                & 141         & 1,447    & 23,446     \\
\hline
number of RDF triples & 253,566,996 & \begin{tabular}[c]{@{}l@{}}27,258,977\\ (4,655,416,683)\end{tabular} & \begin{tabular}[c]{@{}l@{}}11,137,852\\ (119,684,431)\end{tabular} &  \begin{tabular}[c]{@{}l@{}}348,094,663 \\ (2,489,856,093)\end{tabular} &    48,348,838      & \begin{tabular}[c]{@{}l@{}}48,292,483 \\ (267,990,918)\end{tabular}\\
\hline
number of instances   & 17,653,785  & \begin{tabular}[c]{@{}l@{}}1,323,452 \\ (95,312,952)\end{tabular}  & \begin{tabular}[c]{@{}l@{}}287,752\\ (7,362,499)\end{tabular}      &  \begin{tabular}[c]{@{}l@{}}19,707,176\\ (73,260,077)\end{tabular}   &    1,390,438      &  \begin{tabular}[c]{@{}l@{}}33,535,913\\ (115,880,746)\end{tabular}  \\
\hline
\end{tabularx}
\end{adjustbox}
\caption{\label{basic-stats} basic statistics(target language:korean, figures in parentheses are statistics for all language data)} 
\end{table*}

Raftel, introduced in this work, is a knowledge graph that integrates Wikidata and Naver databases on various domains. Based on the class structure and properties of Wikidata, an active and fast-generating knowledge graph based on the community, the basic ontology was designed, classes were added and removed, hierarchical relationships were adjusted, and attributes were added and removed to alleviate the complexity of the ontology and increase consistency.

In addition, since Raftel is a knowledge graph generated based on Korean data, other knowledge graphs were filtered based on the entity with the Korean label. Since Google Knowledge graph can query API based on specific keywords, data with Korean labels that exist in Wikidata were imported. 

Ontology data were refined differently depending on whether or not ontology data is provided with a separate file for each knowledge graph. For DBpedia and YAGO, we used the ontology file provided by themselves. In the case of YAGO, only the classes defined in the ontology file were included for the calculation because the number of classes defined in the ontology file and the number of classes corresponding to B of the RDF triple  \textit{`A - instance of - B'}  was different. The ontology of Google Knowledge Graph is based on schema.org, so we used \textit{schema.org}'s ontology data for Google KG.\footnote{https://developers.google.com/knowledge-graph} For Wikidata and Freebase, where ontology data is not provided with a separate file, we established an ontology by refining the knowledge graph RDF triple data. Wikidata's ontology was extracted using the \textit{`subclass of'} property. In the case of properties for each class, the properties used in the class were collected by mapping the class to the instance of the RDF triple. Freebase's class was targeted at RDF triple's subject with a property of \textit{`<http://www.w3.org/1999/02/22-rdf-syntax-ns\#type>'} and an object of \textit{`<http://www.w3.org/2000/01/rdf-schema\#Class>}. In addition, Freebase does not specify a hierarchy between classes, so there is no root class (\citet{Chah2018OKGW}). To calculate metrics, we created a root class and connected all classes as subclasses of the root class. The property was targeted at RDF triple's subject whose object ends with `Property>' among objects with a property of \textit{`<http://www.w3.org/1999/02/22-rdf-syntax-ns\#type>'}, and only those with a property of \textit{`<http://www.w3.org/2000/01/rdf-schema\#domain>'} were explicitly provided. (Table~\ref{freebase-preperation})

\subsection{Basic Statistics}

Table ~\ref{basic-stats} shows the basic statistics for knowledge graphs. The number of classes and properties was calculated for ontology, and the number of RDF triple and instances were calculated for knowledge graph RDF triple data. It includes an RDF triple with instances of classes or properties that were not defined in the ontology. In the case of ontology, all classes and properties were considered regardless of the Korean label.  Other analysis include RDF triples of the instance with a Korean label among all RDF triples of the knowledge graph. Numbers in parentheses are analytical values for the entire language. For Google KG, since we  extracted data that exist in Raftel, analysis of the entire language was not conducted separately.

In terms of classes and properties that construct the ontology, Wikidata and Freebase have more than 50,000 classes, which is about 200 times more than Raftel and YAGO, which have fewer classes. In the case of properties, DBpedia and Freebase have more than 20,000 properties, 100 times more than YAGO, which has the least properties, and 30 times more than Raftel, which has the second least properties.

On the other hand, Raftel and YAGO have relatively large amount of RDF triples and instances. Compared to DBpedia, which has the lowest RDF triple count, the number of RDF triples of YAGO is more than 30 times higher, and in the case of Raftel, it is more than 20 times higher. In the case of the number of instances, Freebase is the largest, but for Freebase, the number of RDF triples is small because most RDF triples are composed of  \textit{`instance of'} relationship. It can be seen from the fact that the number of an RDF triple is about 1.4 times the number of instances. Next, the number of YAGO and Raftel instances is 190 million and 170 million, respectively, more than 60 times that of DBpedia, which has the lowest number of instances. For YAGO, there are about 18.46 million instances belonging to the \textit{`scholarArticle'} class, accounting for 97\% of the total, and RDF triples with the \textit{`scholarArticle'} instance as the subject account for more than 90\% (312, 203, 867).\\

\section{Structural Quality Metrics}

\subsection{What is a good Knowledge Graph?}  \label{Good Knowledge Base}

A good knowledge graph should have a fine-grained ontology structure that can precisely express information in the real world, and instances and triples should make full use of the ontology's classes and properties. By categorizing this perspective into the four categories listed below, a structural quality metrics that can quantify each content was developed.
 
First, class hierarchy must be abundantly subdivided in the ontology. Taking the \textit{`Person' class} as an example, the more the \textit{Person} is divided into horizontals like \textit{`Artist, Athlete, Politician, Doctor'}, and the more split into verticals like \textit{`Person → Artist → Musician'}, the better the ontology. Compared to the case where only \textit{`Person' class} exists for the class related to people, if it is divided into more classes according to occupation, ontology would be more powerful in various tasks by narrowing the scope of the classification of entities. For example, in the entity disambiguation task, the range to which the entity of the person with the same name belongs is specified, making it easy to distinguish. If the \textit{`Musician'} and \textit{`Author'} classes are added as subclass to \textit{`Artist'}, which is a subclass of \textit{`Person'}, the ontology will provide more specific information to application tasks.

Second, when it is subdivided from superclass to subclass in ontology, the more properties that are not in superclass in subclass, the better. In the case of vertically splitting the class, the number of properties that the subclass has should increase. For example, when \textit{Athlete} class is a subclass of \textit{Person} class, the \textit{Person} class has universal properties such as \textit{"parent"}, \textit{"birth date"}, and \textit{"birth place"}. The more properties (e.g. \textit{"back number"}, \textit{"team"}, \textit{"world ranking"}, \textit{"league"}, etc.) the \textit{Athele} has, the more specific quality of information that can be obtained as a class is added.

Third, classes and properties defined in the ontology must be used sufficiently in the knowledge graph. Even if classes and properties are defined in detail, ontologies are only useful if they are applied to the knowledge graph. The \textit{Person} class can be divided into classes like "Chef of Chinese Restaurant in Seoul" or "4th grade music teacher of the elementary school". Nevertheless, if the number of instances is too small relative to the whole number of instances in the knowledge graph, it is difficult to say that adding the class is beneficial. Likewise, specific properties such as "number of debut songs sung at concerts" and "administrative district where the most fans live" can be defined for the \textit{Musician} class, but if data does not exist and is not used as actual RDF triples, it is better not to add it.

Fourth, though the quality increases as the class is subdivided, it is negative if the complexity increases in this process. Multiple inheritance is an example of a factor that describes ontology complexity. Multiple inheritance means that one class has several superclasses. For example, the \textit{Hospital} class is a subclass of \textit{`Facility'}, a space that provides a specific function, and a subclass of \textit{`Organization'}, a group of employees including doctors and nurses.  Avoiding multiple inheritance is preferable, unless it is necessary, like in the case of the \textit{Hospital} where both location information, which is a facility characteristic, and member information, which is an organization characteristic, are crucial. This is because when subclass and superclass are connected in a many-to-many relationship, the complexity of understanding and utilization of ontology increases.

\begin{figure*} 
  \includegraphics[width=\textwidth ]{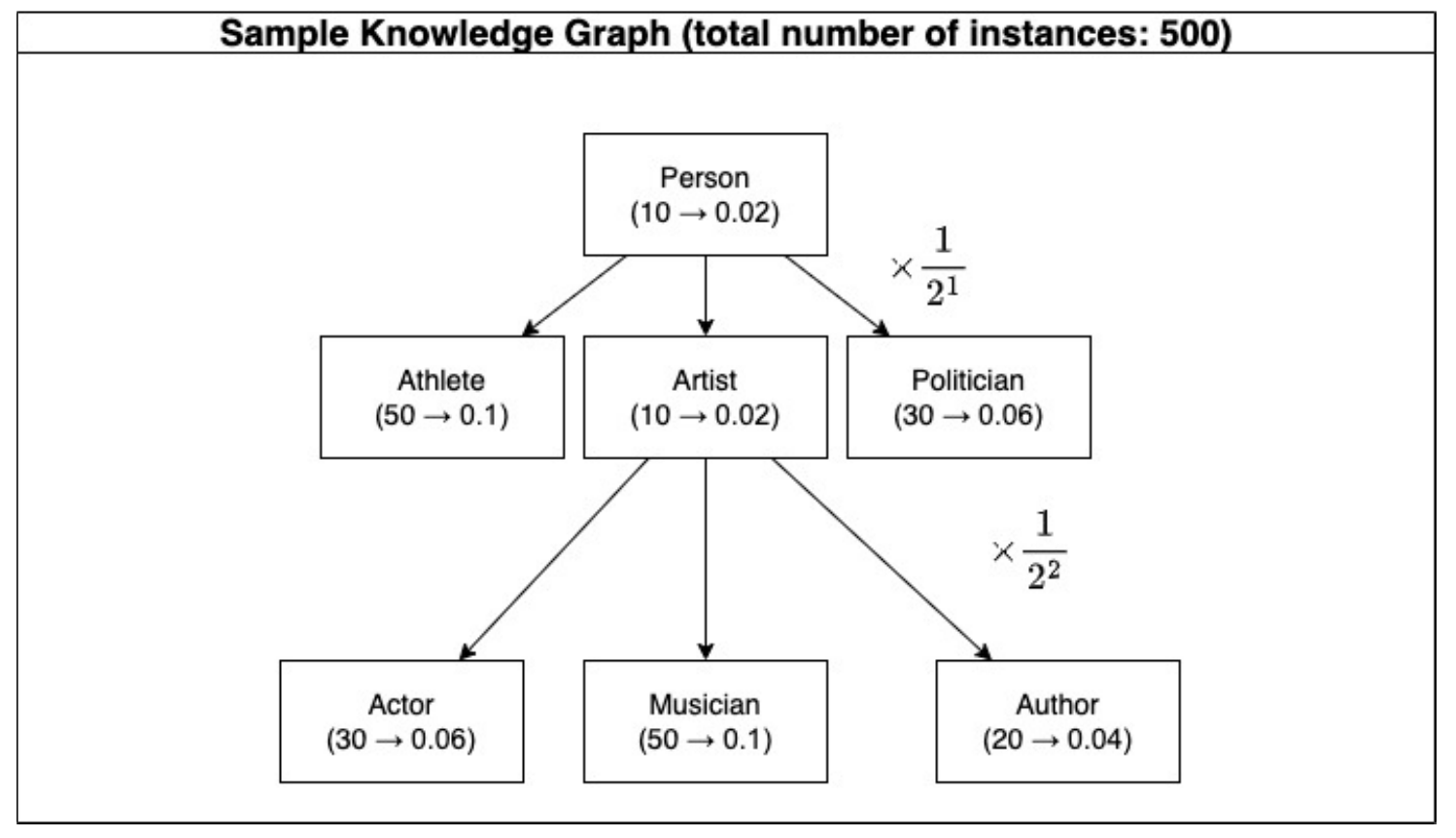} 
  \caption{ \label{Class Instantiation Example} Class Instantiation Example}
\end{figure*}

\subsection{Structural Quality Metrics}

We present a structural quality metric that can measure the quality of good knowledge graphs presented by ~\ref{Good Knowledge Base}. ~\ref{Instantiated Class Ratio}, ~\ref{instantiated Property Ratio} have been examined in previous studies, and ~\ref{Class Instantiation}, ~\ref{Subclass Property Acquisition}, ~\ref{Subclass Property Instantiation}, ~\ref{Inverse Multiple Inheritance}, are newly introduced in this work.

\subsubsection{Instantiated Class Ratio} \label{Instantiated Class Ratio}

\textit{Instantiated Class Ratio} refers to the ratio of classes with instances among classes defined in the ontology. It is an indicator of how well the class of ontology is actually being used. In obtaining \textit{Instantiated Class Ratio} for ontology (~\ref{instantiated class ratio formula}), $N(C)$ means the total number of classes in the Ontology, and $N(IC)$ means the number of classes in which instances exist.

\begin{equation} \label{instantiated class ratio formula}
ICR(Ontology) =  {N(IC) \over N(C)}
\end{equation}

\subsubsection{Instantiated Property Ratio}  \label{instantiated Property Ratio}

\textit{Instantiated Property Ratio} refers to the ratio of properties actually used in RDF triple among the properties defined in the ontology. It is an indicator of how well the properties of the ontology are actually being used. In obtaining \textit{Instantiated Property Ratio} for ontology (~\ref{instantiated Property Ratio formula}), $N(P)$ denotes the total number of properties of the Ontology, and $N(IP)$ denotes the number of properties used in RDF triples.

\begin{equation} \label{instantiated Property Ratio formula}
IPR(Ontology) = {N(IP) \over N(P)}
\end{equation}

\subsubsection{Class Instantiation}  \label{Class Instantiation}

\textit{Class Instantiation} is a metric that assesses how much in detail classes are defined in the ontology and how much they are actually instantiated. For each class included in the knowledge graph, the class instantiation is calculated and summed to be used as an indicator representing the knowledge graph. In a formula (~\ref{class intantiation formula}) to obtain \textit{Class Instantiation} for a particular \textit{Class}, $n_c$ means the number of subclasses that the \textit{Class} has, $ir(c)$ means instantiated ratio, which is \textit{`number of instances of the Class / number of all instances in knolwedge graph'}, $c_i$ is the i-th subclass the \textit{Class} has, $d$ means the distance between the \textit{Class} and $c_i$  .

\begin{equation} \label{class intantiation formula}
CI(Class) = \sum_{i = 1}^{n_c}{ir(c_i)\over2^{d(c_i)}}
\end{equation}

The process of calculating \textit{Class Instantiation} for \textit{`Person'} class in a knowledge graph such as Figure ~\ref{Class Instantiation Example} is as follows. The total number of instances of the knowledge graph is 500, of which the number of instances of \textit{`Person'} and \textit{Person}'s subclass is 200. For all subclasses under the \textit{`Person'} class, the proportion of the class's instance to the total instance is calculated. Let's say this is a "weight". For example, to calculate the weight for \textit{`Artist'}, use the number of direct instances of the \textit{Artist}, not the instances of \textit{`Actor'}, \textit{`Musician'}, or \textit{`Author'}. \textit{`Ariana Grande'} instance is not a direct instance of the \textit{`Artist'} because the singer is an instance of the \textit{Artist}'s subclass, \textit{`Musician'}. Since \textit{`Pablo Picasso'} instance does not belong to the \textit{`Artist'}'s subclasses , it becomes a direct instance of the \textit{`Artist'}. In the figure, the class is represented with rectangle box, and in the box, \textit{(number of instances → weights)} is denoted.

\textit{Person}'s \textit{Class Instantiation} accumulates weights from the subclass farthest from \textit{`Person'}. Weights are not added as they are, but divided by $2^ {Depth From Person Class}$. The farther away from the \textit{`Person'} class, the more penalty the class's weight has. As a result, \textit{Class Instantiation}  of \textit{`Person'} is calculated as  $0.1 + {(0.02 + 0.01 + 0.06) \times {1\over{2^1}}}  + {(0.06 + 0.1 + 0.04) \times{ 1\over{2^2}}}$.

\textit{Class Instantiation} is obtained in the same way as above for all the classes including \textit{`Artist'}, \textit{`Musician'}, and \textit{`Creative Work'} that exist in the ontology.

For \textit{Class Instantiation}, the more classes are divided, the more each class is fully utilized, and the higher the weight, the greater the value of \textit{Class Instantiation}. In addition, the penalty according to depth was applied to prevent the side effects of increasing the score as the class is subdivided into verticals unconditionally.

\subsubsection{Subclass Property Acquisition}  \label{Subclass Property Acquisition}

\textit{Subclass Property Acquisition} is a metric that measures how many properties are defined in the subclass that is not in the superclass in the ontology. For example, if the \textit{`Person'} class is a subclass of the \textit{`Entity'} class, properties  that are not defined in the \textit{`Entity'} class such as `\textit{children}', '\textit{academic degree}', and '\textit{spouse}', can be added. Furthermore, if the \textit{`Actor'} class is a subclass of the \textit{`Person'} class, properties like `\textit{character role}', `\textit{cast member of}' can be added.  The \textit{Subclass Property Acquisition} is the average value obtained by number of newly added properties that are not in the superclass for all classes of the ontology, except for the root class (e.g., \textit{Entity} class). 

In the formula for obtaining \textit{Subclass Property Acquisition} for Ontology (~\ref{Subclass Property Acquisition formula}), $P$ denotes property set and $N(P)$ is a function for the number of elements in the property set. For all `superclass-subclass' relationships present in Ontology, $N(P_{subclass} - P_{superclass})$ is calculated, and the number of properties present in the subclass is obtained and summed. For all classes in Ontology, \textit{Subclass Property Acquisition} is calculated and averaged by $N(C)$, the number of classes.

\begin{equation} \label{Subclass Property Acquisition formula}
SPA(Ontology) = {{\sum(N_i(P_{sublass} - P_{superclass}))} \over N(C)}
\end{equation}

\begin{table*}[ht!]
\renewcommand{\arraystretch}{1.4}
\renewcommand{\tabcolsep}{3.5mm}
\centering
\begin{adjustbox}{width=1\textwidth}
\small
\begin{tabularx}{\textwidth}{lllllll}
\hline
                                & Raftel & Wikidata & DBpedia                                                 & YAGO   & Google KG             & Freebase              \\
                                \hline
Instantiated Class Ratio        & 0.941   &  \begin{tabular}[c]{@{}c@{}}0.004 \\ (0.334)\end{tabular}    & \begin{tabular}[c]{@{}c@{}}0.470\\ (0.540)\end{tabular}   & \begin{tabular}[c]{@{}c@{}} 0.820 \\ (0.966)\end{tabular}  & 0.099 &         \begin{tabular}[c]{@{}c@{}}0.046 \\ (0.314)\end{tabular}           \\
\hline
Instantiated Property Ratio     &  1   &  \begin{tabular}[c]{@{}c@{}}1 \\ (1)\end{tabular}       &                           \begin{tabular}[c]{@{}c@{}} 0.99 \\ (1)\end{tabular}                             &  \begin{tabular}[c]{@{}c@{}} 0.90  \\ (0.96)\end{tabular}      & 1                     &  \begin{tabular}[c]{@{}c@{}} 0.002    \\ (0.003)\end{tabular}                 \\
\hline
Class Instantiation             & 0.941  &  \begin{tabular}[c]{@{}c@{}}0.716  \\ (0.743)\end{tabular}    & \begin{tabular}[c]{@{}c@{}}0.900\\ (0.949)\end{tabular} &  \begin{tabular}[c]{@{}c@{}}0.886 \\ (0.616)\end{tabular} & 0.660 &     \begin{tabular}[c]{@{}c@{}} 0.874   \\ (0.749)\end{tabular}           \\
\hline
Inverse Multiple Inheritance    & 0.975  & 0.962    & 0.971                                                   & 0.942  & 0.952                 & 1                     \\
\hline
Subclass Property Acquisition   & 6.54   & 40.94    & 63.57                                                   & 2.23   & 0.0                     & 1                     \\
\hline
Subclass Property Instantiation & 0.0857 &  \begin{tabular}[c]{@{}c@{}} 0.0133    \\ (0.00001)\end{tabular}     &   \begin{tabular}[c]{@{}c@{}} 0.0841   \\ (0.0668)\end{tabular}                                                & \begin{tabular}[c]{@{}c@{}}  0.0003    \\ (0.00001)\end{tabular}   & 0.0                     & \begin{tabular}[c]{@{}c@{}} 0.0    \\ (0.0)\end{tabular} \\
\hline
\end{tabularx}
\end{adjustbox}
\caption{\label{structural-quality} structural quality metric evaluation (figures in parentheses are statistics for all language data)} 
\end{table*}

\subsubsection{Subclass Property Instantiation}  \label{Subclass Property Instantiation}

\textit{Subclass Property Instantiation} quantifies how much the properties are used in the RDF triples when the properties of the subclass that are not in the superclass are defined in the ontology. For example, if an \textit{`Actor'} class adds `\textit{cast member of}' and `\textit{character role}' property that are not in the superclass \textit{`Person'} class, the more unique properties of the actor are used in RDF triples, such as \textit{"Tom Cruise - cast member of - Mission Impossible"} and \textit{"Tom Cruise - character role - Ethan Hunt"}, the better structure knowledge graph has. Knowledge graph's \textit{Subclass Property Instantiation} is average of \textit{Subclass Property Instantiation} of all the classes.

In the formula for obtaining \textit{Subclass Property Instantiation} for a particular \textit{Class} (~\ref{Subclass Property Instantiation formula}), $T$ is a set of RDF triples, and N(T) is a function of obtaining the number of RDF triples. $N(T_{class} - T_{class\_superclass})$ is the number of RDF triples of \textit{Class} excluding RDF tripels which uses predicates defined for superclass.

\begin{equation} \label{Subclass Property Instantiation formula}
SPI(Class) = {N(T_{Class} - T_{class\_superclass}) \over N(T_{Class})}
\end{equation}

To compute a \textit{Subclass Property Instantiation} for an \textit{`Actor'} class, first, count the number of all triples with an \textit{`Actor'} class's instance as the subject. In addition to RDF triples like \textit{"Tom Cruise - cast memeber of - Mission Impossible"} and \textit{"Tom Cruise - character role - Ethan Hunt"} , count all the RDF triples including \textit{"Tom Cruise - Birth Place - Syracuse"}, \textit{"Tom Cruise - Nationality - United States"}, and \textit{"Tom Cruise - Name - Tom Cruise"}. This becomes denominator of the \textit{Subclass Property Instantiation.} Next, count the number of triples in which the properties added in the \textit{`Actor'} class are used. Except for \textit{"Tom Cruise - Birth Place - Syracuse"}, \textit{"Tom Cruise - Nationality - United States"}, and \textit{"Tom Cruise - Name - Tom Cruise"}, only RDF triples such as \textit{"Tom Cruise - cast member of - Mission Impossible"} and \textit{"Tom Cruise - character role - Ethan Hunt"} are considered. This is the numerator of  \textit{Subclass Property Instantiation}. By dividing the number of RDF triples used by the property added in the \textit{`Actor'} by the total number of triples in the \textit{`Actor'}, we can see how the unique property increases in RDF triple as the \textit{`Actor'} was subdivided from \textit{`Person'}. For example, if the \textit{Actor}'s \textit{Subclass Property Instantiation} is 0.05, it means that the RDF triple of \textit{`Actor'} with new properties has increased by 5\% compared to the superclass \textit{`Person'}.

\subsubsection{Inverse Multiple Inheritance}  \label{Inverse Multiple Inheritance}

\textit{Inverse Multiple Inheritance} evaluates the simplicity of the knowledge graph. If multiple inheritance occurs frequently  in which a single class has numerous superclasses, might make it challenging to use the knowledge graph because of the complexity of the class relationship. Inverse multiple inheritance was devised to measure how little multiple inheritance appears. The average number of superclasses per class is computed to obtain the average multiple inheritance, and take the reciprocal of it. Therefore, the higher the \textit {Inverse Multiple Inheritance}, the simpler the knowledge graph is. In (~\ref{Inverse Multiple Inheritance formula}), $N_c$ represents the total number of classes in the ontology, and $C_i$ represents each class in the ontology, $nsup(C)$ represents the number of direct superclasses in the class.

\begin{equation} \label{Inverse Multiple Inheritance formula}
IMI(Ontology) ={1\over{\sum_{i = 1}^{N_c}{nsup(C_i)} \over N_c}}
\end{equation}

The six structural quality metrics determine whether knowledge graph can express knowledge abundantly through a detailed ontology. Among them, \textit{Class Instantiation} and \textit{Subclass Property Instantiation} have the characteristics of a comprehensive indicator that can reflect classes or attribute's degree of subdivision and actual utilization.

\subsection{Structural Quality Metric Evaluation Result}

Table ~\ref{structural-quality} is the analysis of structural quality metric with five knowledge graphs on the web and Raftel. First, looking at the metric related to the degree of segmentation and usage of classes, YAGO and Raftel have the smallest number of classes in Table ~\ref{basic-stats}, but more than 80\% of classes were instantiated. On the other hand, Wikidata, Freebase, and Google KG instantiated less than 10\% compared to the large number of classes defined in Ontology. According to the results of the \textit{Class Instantiation} analysis, DBpedia and Raftel fully utilize fine-grained ontology in the knowledge graph. When comparing DBpedia and YAGO, even though YAGO has higher \textit {Instantiated Class Ratio}, DBpedia's  \textit{Class Instantiation} shows that it is divided into vertical and horizontal classes, and the classes are actively used in the knowledge graph. In the case of Google KG, the low number of classes that could be imported through the API was reflected in the low \textit{Instantiated Class Ratio} and \textit{Class Instantiation}. Freebase does not define the hierarchy between classes, so it seems to affect the low value of \textit {Class Instantiation}. When it comes to \textit{Inverse Multiple Inheritance}, Freebase is calculated as 1 because classes do not have parent classes, and has the most concise ontology structure. Schema.org(Google KG's ontology) and YAGO have a high complexity due to its relatively frequent multiple inheritance.

Referring the property-related metrics, DBpedia and Wikidata have a large number of properties defined in the ontology(Table ~\ref{basic-stats}) and \textit{Subclass Property Acquisition} shows the large number of properties are added as classes are subdivided. On the other hand, for Freebase, the number of properties of Table ~\ref{basic-stats} is large, but the value of \textit{Subclass Property Acquisition} is low because the ontology has no class hierarchy. Google KG only provides basic information about `name, image, description, class' through the API, so \textit {Subclass Property Acquisition} appears low. Looking at \textit{Subclass Property Instantiation}, Wikidata has richly defined properties according to class segmentation in the ontology, but the degree of use is relatively low. DBpedia has abundant properties and those properties are well used in RDF triple. YAGO has small number of properties(Table ~\ref{basic-stats}). Also, \textit{Subclass Property Acquisition} and \textit{Subclass Property Instantiation} infers that the degree of segmentation is low.

Raftel, which is based on Wikidata ontology, appears to have high scores in \textit{Class Instantiation} and \textit{Subclass Property Instantiation}, which are the comprehensive score, by organizing classes and properties abundantly in data while refined them according to the criteria of ~\ref{Good Knowledge Base}.

\begin{table*}[ht!]
\renewcommand{\arraystretch}{1.4}
\renewcommand{\tabcolsep}{3.5mm}
\centering
\begin{adjustbox}{width=1\textwidth}
\small
\begin{tabularx}{\textwidth}{lllllll}
\hline
                                & Raftel & Wikidata & DBpedia                                                 & YAGO   & Google KG             & Freebase              \\
                                \hline
$0.0\times CM+1.0\times PM$        & 7.31   & 6.40    & \textbf{9.91}   &  3.81   & 4.00 & 1.04                  \\
\hline
$0.25\times CM+0.75\times PM$      & 7.66      & 5.47        & \textbf{9.07}                                                    & 4.37   & 3.45                    & 2.39                  \\
\hline
$0.5\times CM+0.5\times PM$              & \textbf{8.01}  & 4.51    & 8.23 & 4.92  & 2.90 & 3.72                 \\
\hline
$0.75\times CM+0.25\times PM$     & \textbf{8.36}  & 3.57    & 7.40                                                   & 5.47  & 2.34                 & 5.06                     \\
\hline
$1.0\times CM+0.0\times PM$    & \textbf{8.71}  & 2.63    & 6.56                                                   & 6.03   & 1.79                    & 6.4                    \\
\hline
\end{tabularx}
\end{adjustbox}
\caption{\label{structural-quality-normalize} normalization of structural quality metric (target language:korean), CM = Class Metrics, PM = Property Metrics} 
\end{table*}

Table ~\ref{structural-quality-normalize} shows comprehensive analysis of structural quality metric by categorizing metrics to \textit{Class Metric} (CM) including \textit{Instantiated Class Ratio}, \textit{Class Instantiation}, \textit{Inverse Multiple Inheritance} and \textit{Property Metric} (PM) including \textit{Instantiated Property Ratio}, \textit{Subclass Property Acquisition}, \textit{Subclass Property Instantiation} and calcuating weighted average of them.

Each metric of the structural quality metric was normalized to have a minimum value of 1 and a maximum value of 10, and then metrics belonging to \textit{Class Metric} and metrics belonging to \textit{Property Metric} were averaged to obtain a representative value. After that, the characteristics of the knowledge graph were examined by varying the weights of CM and PM. When the proportion of PM is large, DBpedia showed the highest score, and when the weighted average was the same or the proportion of CM was larger, Raftel showed the highest score. Through this, if the degree of segmentation of the property is important, the quality of DBpedia can be judged to be high, and if the degree of segmentation of the class is important, the quality of Raftel can be judged to be high.

\section{Conclusion}

In this study, six structural quality metrics were proposed as indicators to evaluate the quality of knowledge graphs. Reflecting the view that `Knowledge graphs should have the ontology that can express knowledge in the real world, and the knowledge graph RDF triples should utilize the ontology sufficiently', we present the \textit{Instantiated Class Ratio}, \textit{Instantiated Property Ratio}, \textit{Class Instantiation}, \textit{Subclass Property Acquisition}, and \textit{Subclass Property Instantiation}. Also, \textit{Inverse Multiple Inheritance} was introduced to ease the complexity of the ontology. 

In addition, the structural quality metric was applied to five cross-domain knowledge graphs on the web and Naver's integrated knowledge graph, Raftel for the comparative analysis. Compared to the structure evaluation conducted only in terms of the size and distribution of graphs, it was able to gain in-depth insights on the quality of knowledge graphs.

Structural quality metric sees `structure' as an important factor in determining the quality of knowledge graphs. According to the results of the structural quality metric analysis, some knowledge graphs with many classes and properties in their ontology have low degree of segmentation and instantiation. On the contrary, some knowledge graphs that have less classes and properties compared to others described knowledge in detail with specified classes and their distinct characteristics. Of course, since each knowledge graph has a different orientation, knowledge graphs with a low score in the structural quality metric can also show good scores in the quality metric in different dimensions. In future studies, it is expected that the strengths and weaknesses of each knowledge graph to be examined with multi-dimensional point of view by applying various metrics from the data quality perspective presented in previous works.

\bibliography{custom}

\begin{thebibliography}{27}
\expandafter\ifx\csname natexlab\endcsname\relax\def\natexlab#1{#1}\fi

\bibitem[{Auer et~al.(2007)Auer, Bizer, Kobilarov, Lehmann, Cyganiak, and
  Ives}]{inproceedings}
Sören Auer, Christian Bizer, Georgi Kobilarov, Jens Lehmann, Richard Cyganiak,
  and Zachary Ives. 2007.
\newblock \href {https://doi.org/10.1007/978-3-540-76298-0_52} {Dbpedia: A
  nucleus for a web of open data}.
\newblock volume~6, pages 722--735.

\bibitem[{Bollacker et~al.(2008)Bollacker, Evans, Paritosh, Sturge, and
  Taylor}]{10.1145/1376616.1376746}
Kurt Bollacker, Colin Evans, Praveen Paritosh, Tim Sturge, and Jamie Taylor.
  2008.
\newblock \href {https://doi.org/10.1145/1376616.1376746} {Freebase: A
  collaboratively created graph database for structuring human knowledge}.
\newblock In \emph{Proceedings of the 2008 ACM SIGMOD International Conference
  on Management of Data}, SIGMOD '08, page 1247–1250, New York, NY, USA.
  Association for Computing Machinery.

\bibitem[{Brank et~al.(2005)Brank, Grobelnik, and
  Mladeni{\'c}}]{brank2005survey}
Janez Brank, Marko Grobelnik, and Dunja Mladeni{\'c}. 2005.
\newblock A survey of ontology evaluation techniques.
\newblock In \emph{Proc. of 8th Int. multi-conf. Information Society}, pages
  166--169.

\bibitem[{Chah(2018)}]{Chah2018OKGW}
Niel Chah. 2018.
\newblock Ok google, what is your ontology? or: Exploring freebase
  classification to understand google's knowledge graph.
\newblock \emph{ArXiv}, abs/1805.03885.

\bibitem[{Cheon et~al.(2021)Cheon, Kim, and Kang}]{heeseon}
HeeSeon Cheon, HyunHo Kim, and Inho Kang. 2021.
\newblock Taxonomy induction from wikidata using directed acyclic graph's
  centrality.
\newblock \emph{Human and Language Technology}, 2021.10a:582--587.

\bibitem[{Dong et~al.(2014)Dong, Gabrilovich, Heitz, Horn, Lao, Murphy,
  Strohmann, Sun, and Zhang}]{45634}
Xin~Luna Dong, Evgeniy Gabrilovich, Geremy Heitz, Wilko Horn, Ni~Lao, Kevin
  Murphy, Thomas Strohmann, Shaohua Sun, and Wei Zhang. 2014.
\newblock \href {http://www.cs.cmu.edu/~nlao/publication/2014.kdd.pdf}
  {Knowledge vault: A web-scale approach to probabilistic knowledge fusion}.
\newblock In \emph{The 20th {ACM} {SIGKDD} International Conference on
  Knowledge Discovery and Data Mining, {KDD} '14, New York, NY, {USA} - August
  24 - 27, 2014}, pages 601--610.
\newblock Evgeniy Gabrilovich Wilko Horn Ni Lao Kevin Murphy Thomas Strohmann
  Shaohua Sun Wei Zhang Geremy Heitz.

\bibitem[{F{\"a}rber and Rettinger(2018)}]{Frber2018WhichKG}
Michael F{\"a}rber and Achim Rettinger. 2018.
\newblock Which knowledge graph is best for me?
\newblock \emph{ArXiv}, abs/1809.11099.

\bibitem[{Guo et~al.(2022)Guo, Zhuang, Qin, Zhu, Xie, Xiong, and He}]{9216015}
Qingyu Guo, Fuzhen Zhuang, Chuan Qin, Hengshu Zhu, Xing Xie, Hui Xiong, and
  Qing He. 2022.
\newblock \href {https://doi.org/10.1109/TKDE.2020.3028705} {A survey on
  knowledge graph-based recommender systems}.
\newblock \emph{IEEE Transactions on Knowledge and Data Engineering},
  34(8):3549--3568.

\bibitem[{Heist et~al.(2020)Heist, Hertling, Ringler, and
  Paulheim}]{heist2020knowledge}
Nicolas Heist, Sven Hertling, Daniel Ringler, and Heiko Paulheim. 2020.
\newblock \href {http://arxiv.org/abs/2003.00719} {Knowledge graphs on the web
  -- an overview}.

\bibitem[{Huang et~al.(2019)Huang, Zhang, Li, and Li}]{10.1145/3289600.3290956}
Xiao Huang, Jingyuan Zhang, Dingcheng Li, and Ping Li. 2019.
\newblock \href {https://doi.org/10.1145/3289600.3290956} {Knowledge graph
  embedding based question answering}.
\newblock In \emph{Proceedings of the Twelfth ACM International Conference on
  Web Search and Data Mining}, WSDM '19, page 105–113, New York, NY, USA.
  Association for Computing Machinery.

\bibitem[{Kaffee et~al.(2017)Kaffee, Piscopo, Vougiouklis, Simperl, Carr, and
  Pintscher}]{10.1145/3125433.3125465}
Lucie-Aim\'{e}e Kaffee, Alessandro Piscopo, Pavlos Vougiouklis, Elena Simperl,
  Leslie Carr, and Lydia Pintscher. 2017.
\newblock \href {https://doi.org/10.1145/3125433.3125465} {A glimpse into
  babel: An analysis of multilinguality in wikidata}.
\newblock In \emph{Proceedings of the 13th International Symposium on Open
  Collaboration}, OpenSym '17, New York, NY, USA. Association for Computing
  Machinery.

\bibitem[{Lourdusamy and John(2018)}]{Lourdusamy2018ARO}
Ravi Lourdusamy and Antony John. 2018.
\newblock A review on metrics for ontology evaluation.
\newblock \emph{2018 2nd International Conference on Inventive Systems and
  Control (ICISC)}, pages 1415--1421.

\bibitem[{Nielsen et~al.(2017)Nielsen, Mietchen, and
  Willighagen}]{2f14de77803f4c3fbab040b6b0a63514}
{Finn {\AA}rup} Nielsen, Daniel Mietchen, and Egon Willighagen. 2017.
\newblock \href {https://doi.org/10.1007/978-3-319-70407-4_36} {Scholia,
  scientometrics and wikidata}.
\newblock In \emph{The Semantic Web:}, Lecture Notes in Computer Science, pages
  237--259. Springer Nature Switzerland AG.

\bibitem[{Piscopo and Simperl(2018)}]{10.1145/3274410}
Alessandro Piscopo and Elena Simperl. 2018.
\newblock \href {https://doi.org/10.1145/3274410} {Who models the world?
  collaborative ontology creation and user roles in wikidata}.
\newblock \emph{Proc. ACM Hum.-Comput. Interact.}, 2(CSCW).

\bibitem[{Piscopo and Simperl(2019)}]{10.1145/3306446.3340822}
Alessandro Piscopo and Elena Simperl. 2019.
\newblock \href {https://doi.org/10.1145/3306446.3340822} {What we talk about
  when we talk about wikidata quality: A literature survey}.
\newblock In \emph{Proceedings of the 15th International Symposium on Open
  Collaboration}, OpenSym '19, New York, NY, USA. Association for Computing
  Machinery.

\bibitem[{Prasojo et~al.(2016)Prasojo, Darari, Razniewski, and
  Nutt}]{Prasojo2016ManagingAC}
Radityo~Eko Prasojo, Fariz Darari, Simon Razniewski, and Werner Nutt. 2016.
\newblock Managing and consuming completeness information for wikidata using
  cool-wd.
\newblock In \emph{COLD@ISWC}.

\bibitem[{Raad and Cruz(2015)}]{Raad2015ASO}
Joe Raad and Christophe Cruz. 2015.
\newblock A survey on ontology evaluation methods.
\newblock In \emph{KEOD}.

\bibitem[{Ringler and Paulheim(2017)}]{Ringler2017OneKG}
Daniel Ringler and Heiko Paulheim. 2017.
\newblock One knowledge graph to rule them all? analyzing the differences
  between dbpedia, yago, wikidata \& co.
\newblock In \emph{KI}.

\bibitem[{Spitz et~al.(2016)Spitz, Dixit, Richter, Gertz, and
  Geiß}]{conf/icwsm/SpitzDRGG16}
Andreas Spitz, Vaibhav Dixit, Ludwig Richter, Michael Gertz, and Johanna Geiß.
  2016.
\newblock \href
  {http://dblp.uni-trier.de/db/conf/icwsm/wiki2016.html#SpitzDRGG16} {State of
  the union: A data consumer's perspective on wikidata and its properties for
  the classification and resolution of entities.}
\newblock In \emph{Wiki@ICWSM}, volume WS-16-17 of \emph{AAAI Workshops}. AAAI
  Press.

\bibitem[{Suchanek et~al.(2007)Suchanek, Kasneci, and
  Weikum}]{10.1145/1242572.1242667}
Fabian~M. Suchanek, Gjergji Kasneci, and Gerhard Weikum. 2007.
\newblock \href {https://doi.org/10.1145/1242572.1242667} {Yago: A core of
  semantic knowledge}.
\newblock In \emph{Proceedings of the 16th International Conference on World
  Wide Web}, WWW '07, page 697–706, New York, NY, USA. Association for
  Computing Machinery.

\bibitem[{Tanon et~al.(2016)Tanon, Vrandečić, Schaffert, Steiner, and
  Pintscher}]{44818}
Thomas~Pellissier Tanon, Denny Vrandečić, Sebastian Schaffert, Thomas
  Steiner, and Lydia Pintscher. 2016.
\newblock From freebase to wikidata: The great migration.
\newblock In \emph{World Wide Web Conference}.

\bibitem[{Tartir et~al.(2005)Tartir, Arpinar, Moore, Sheth, and
  Aleman-Meza}]{Tartir2005OntoQAMO}
Samir Tartir, Ismailcem~Budak Arpinar, Michael Moore, A.~Sheth, and Boanerges
  Aleman-Meza. 2005.
\newblock Ontoqa: Metric-based ontology quality analysis.

\bibitem[{Tiddi and Schlobach(2022)}]{TIDDI2022103627}
Ilaria Tiddi and Stefan Schlobach. 2022.
\newblock \href {https://doi.org/https://doi.org/10.1016/j.artint.2021.103627}
  {Knowledge graphs as tools for explainable machine learning: A survey}.
\newblock \emph{Artificial Intelligence}, 302:103627.

\bibitem[{Vrandečić and Krötzsch(2014)}]{42240}
Denny Vrandečić and Markus Krötzsch. 2014.
\newblock \href
  {http://cacm.acm.org/magazines/2014/10/178785-wikidata/fulltext} {Wikidata: A
  free collaborative knowledge base}.
\newblock \emph{Communications of the ACM}, 57:78--85.

\bibitem[{Wang and Strong(1996)}]{10.1080/07421222.1996.11518099}
Richard~Y. Wang and Diane~M. Strong. 1996.
\newblock \href {https://doi.org/10.1080/07421222.1996.11518099} {Beyond
  accuracy: What data quality means to data consumers}.
\newblock \emph{J. Manage. Inf. Syst.}, 12(4):5–33.

\bibitem[{Zaveri et~al.(2018)Zaveri, Kontokostas, Hellmann, Umbrich,
  F\"{a}rber, Bartscherer, Menne, Rettinger, Zaveri, Kontokostas, Hellmann, and
  Umbrich}]{10.3233/SW-170275}
Amrapali Zaveri, Dimitris Kontokostas, Sebastian Hellmann, J\"{u}rgen Umbrich,
  Michael F\"{a}rber, Frederic Bartscherer, Carsten Menne, Achim Rettinger,
  Amrapali Zaveri, Dimitris Kontokostas, Sebastian Hellmann, and J\"{u}rgen
  Umbrich. 2018.
\newblock \href {https://doi.org/10.3233/SW-170275} {Linked data quality of
  dbpedia, freebase, opencyc, wikidata, and yago}.
\newblock \emph{Semant. Web}, 9(1):77–129.

\bibitem[{Zou(2020)}]{Zou_2020}
Xiaohan Zou. 2020.
\newblock \href {https://doi.org/10.1088/1742-6596/1487/1/012016} {A survey on
  application of knowledge graph}.
\newblock \emph{Journal of Physics: Conference Series}, 1487(1):012016.

\end{thebibliography}
\bibliographystyle{acl_natbib}

\onecolumn
\appendix
\section{Appendix}
\label{sec:appendix}

\subsection{Basic Statistics}
\subsubsection{Number of Classes}
\begin{minipage}{\linewidth}
\includegraphics[width=\linewidth]{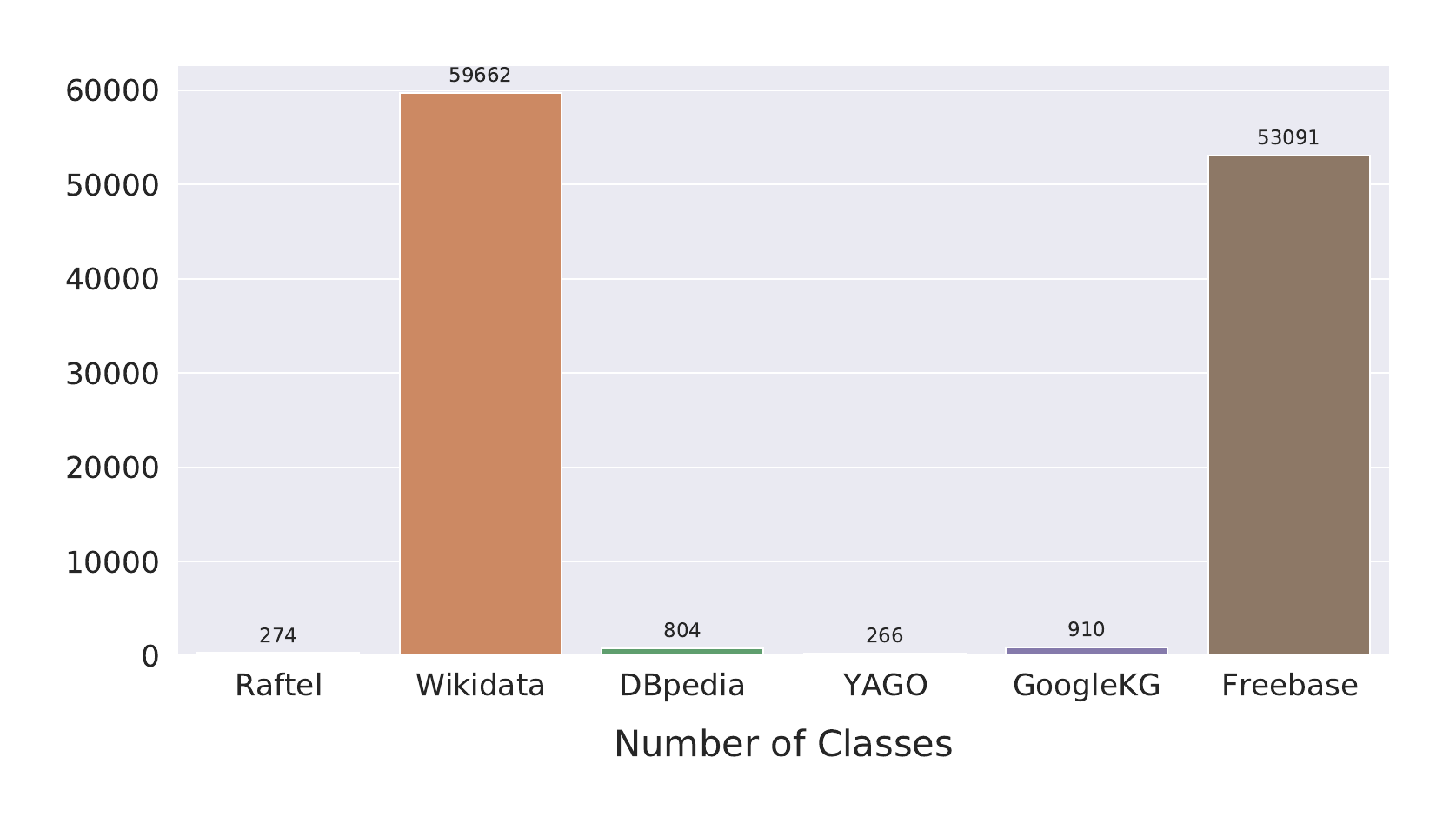}
\captionof{figure}{Number of Classes (target language: Korean)}
\end{minipage}
\subsubsection{Number of Properties}
\begin{minipage}{\linewidth}
\includegraphics[width=\linewidth]{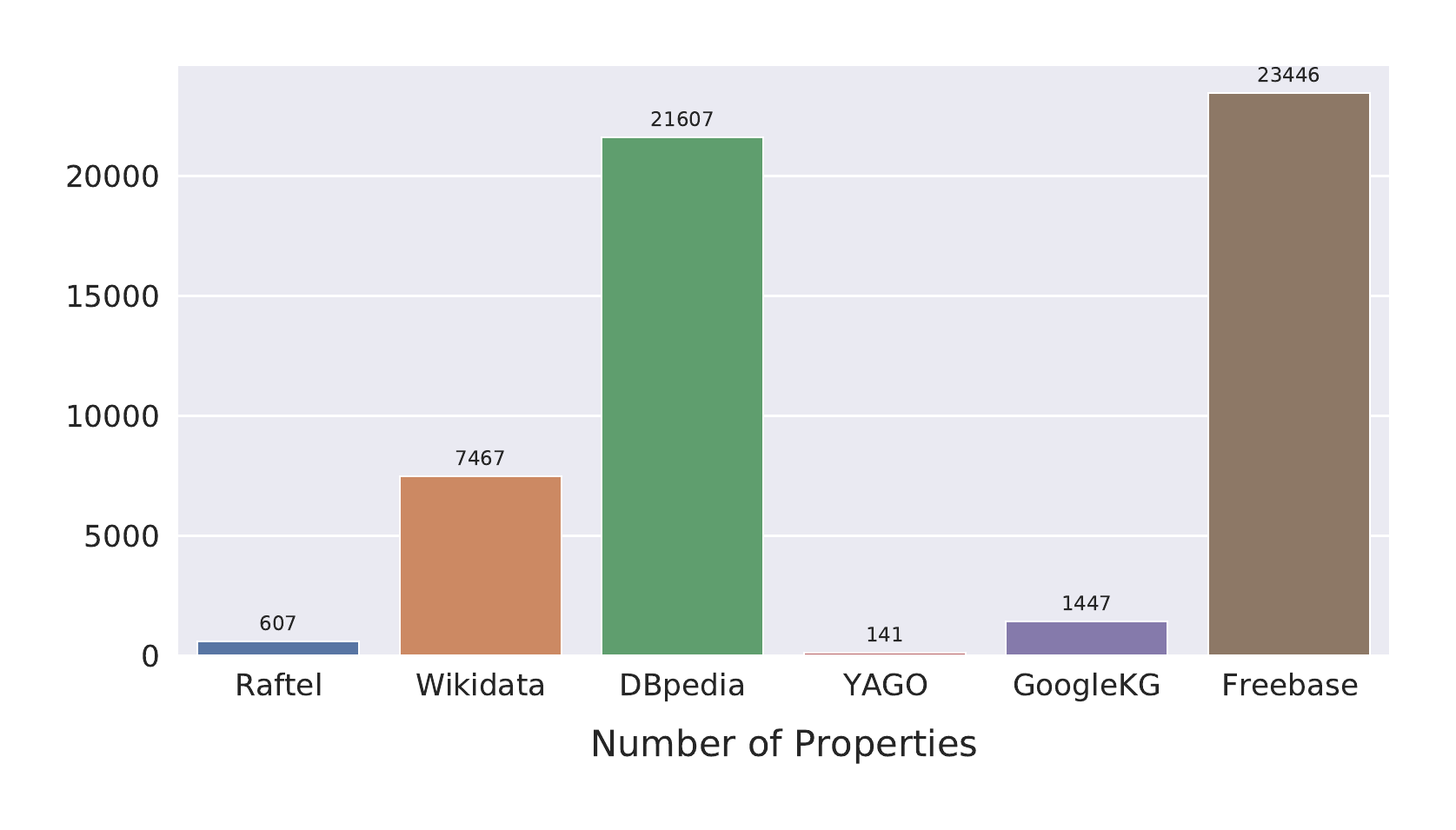}
\captionof{figure}{Number of Properties (target language: Korean)}
\end{minipage}
\subsubsection{Number of RDF Triples}
\begin{minipage}{\linewidth}
\includegraphics[width=\linewidth]{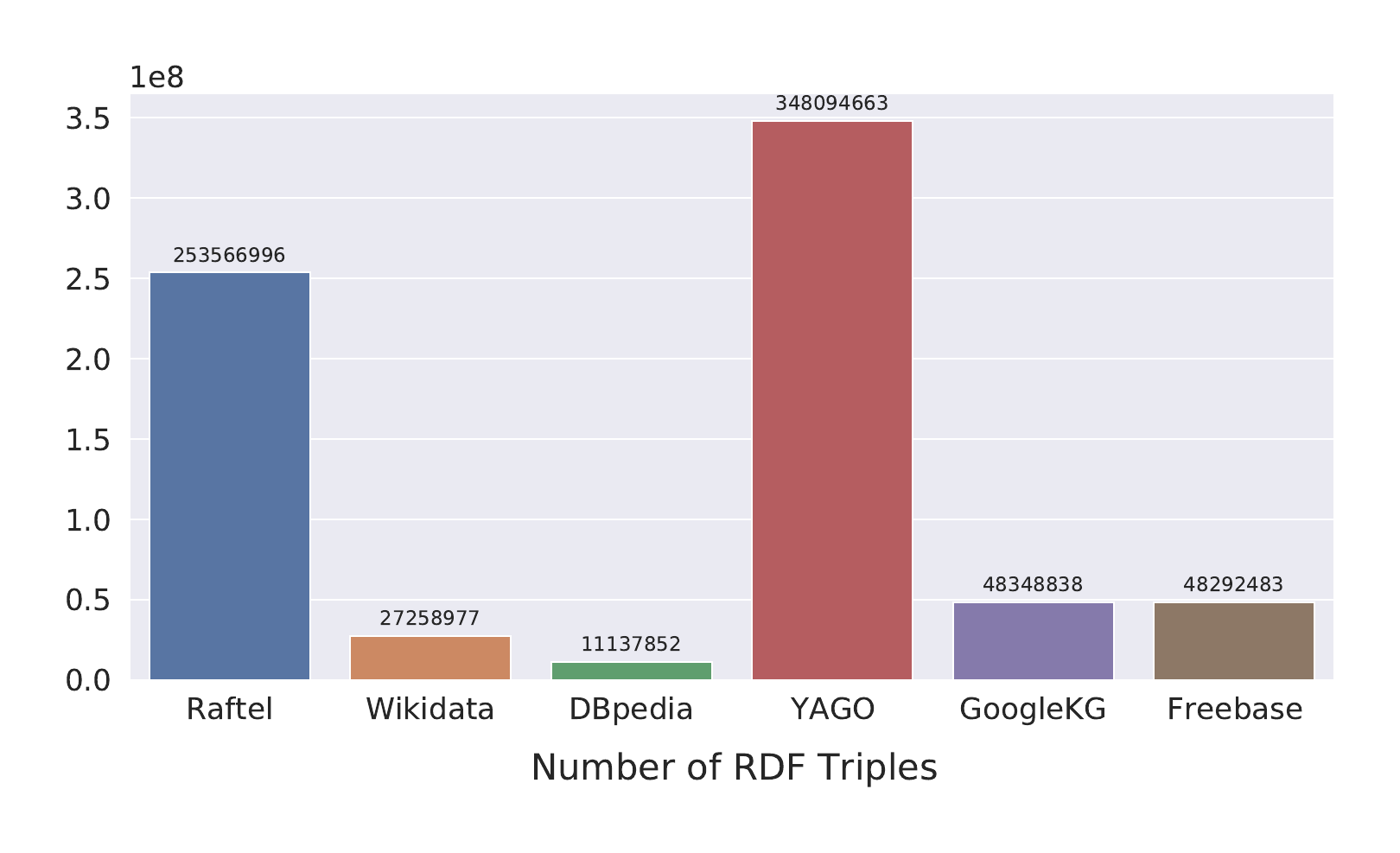}
\captionof{figure}{Number of RDF Triples (target language: Korean)}
\end{minipage}
\subsubsection[width=\textwidth ]{Number of Instances}
\begin{minipage}{\linewidth}
\includegraphics[width=\linewidth]{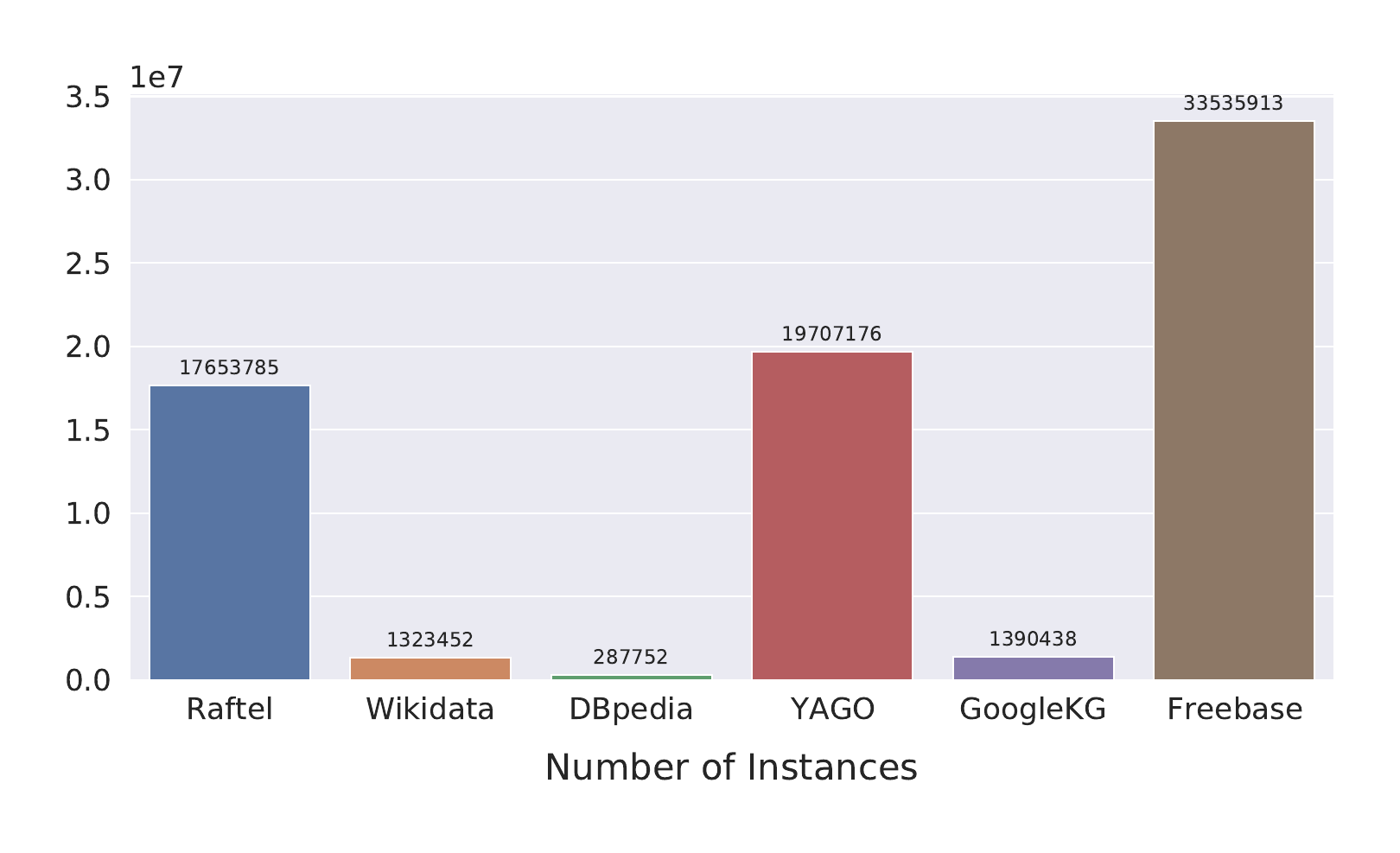}
\captionof{figure}{Number of Instances (target language: Korean)}
\end{minipage}

\subsection{Structural Quality Metrics}
\subsubsection{Instantiated Class Ratio}
\begin{minipage}{\linewidth}
\includegraphics[width=\linewidth]{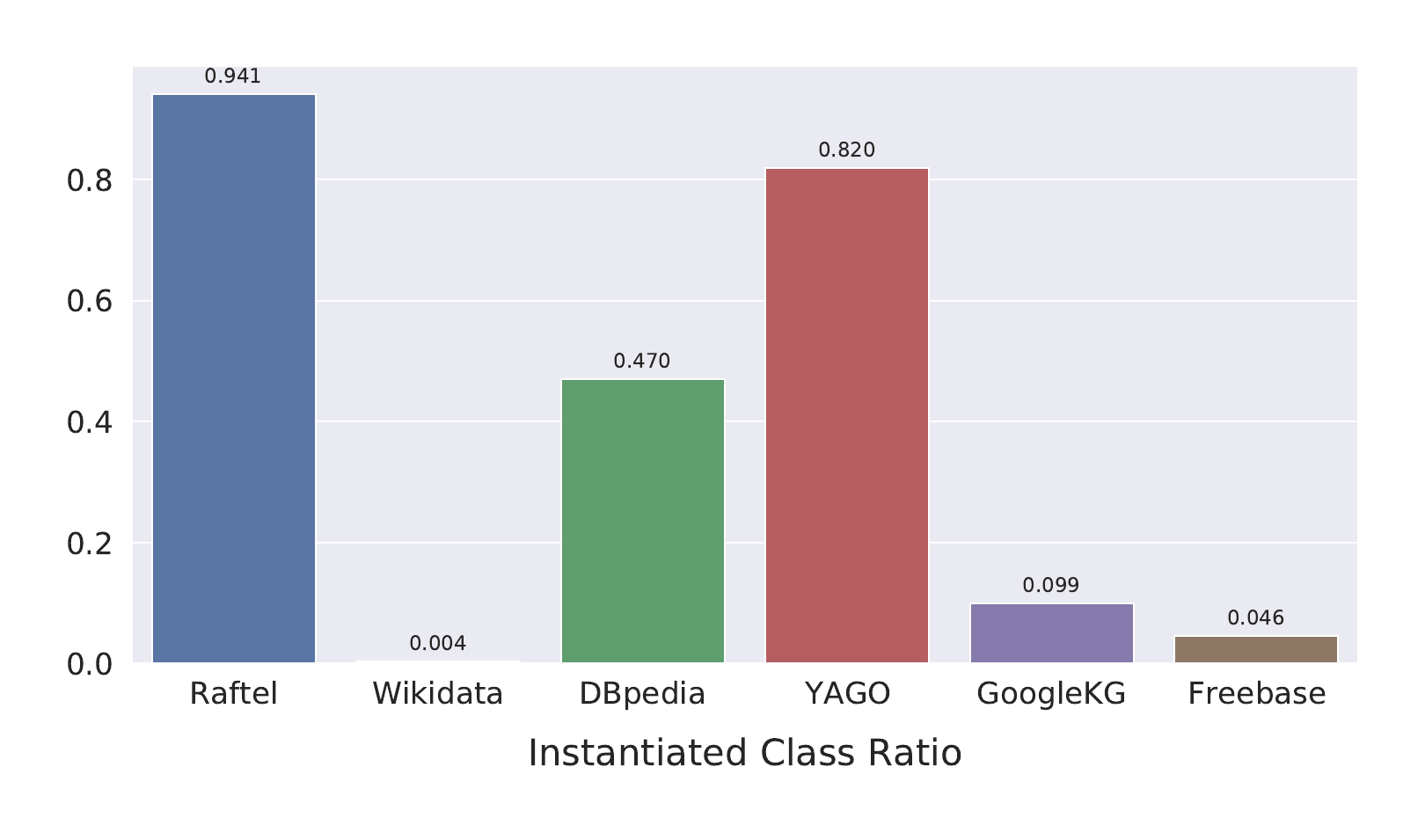}
\captionof{figure}{Instantiated Class Ratio (target language: Korean)}
\end{minipage}
\subsubsection{Instantiated Property Ratio}
\begin{minipage}{\linewidth}
\includegraphics[width=\linewidth]{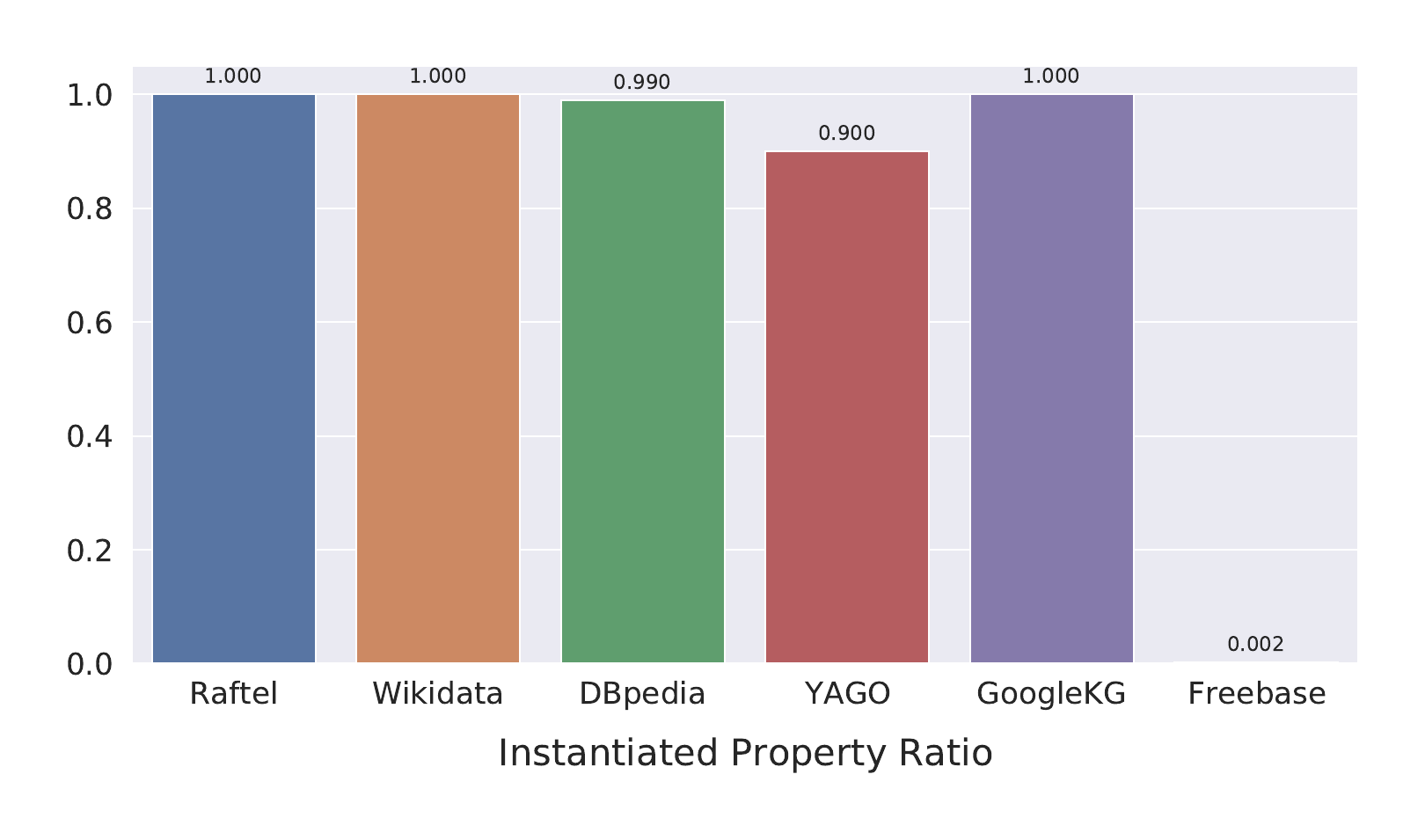}
\captionof{figure}{Instantiated Property Ratio (target language: Korean)}
\end{minipage}
\subsubsection{Class Instantiation}
\begin{minipage}{\linewidth}
\includegraphics[width=\linewidth]{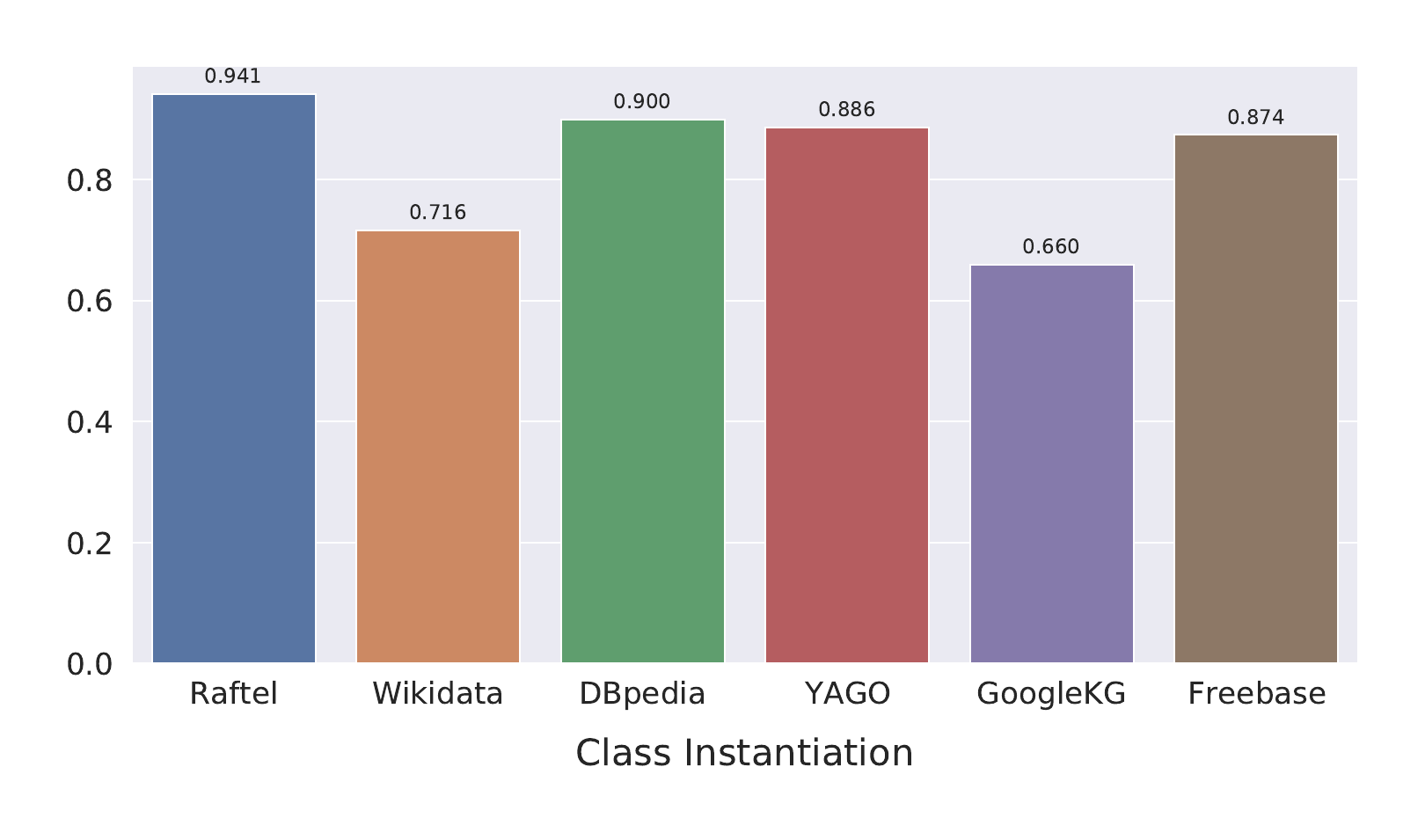}
\captionof{figure}{Class Instantiation (target language: Korean)}
\end{minipage}
\subsubsection{Inverse Multiple Inheritance}
\begin{minipage}{\linewidth}
\includegraphics[width=\linewidth]{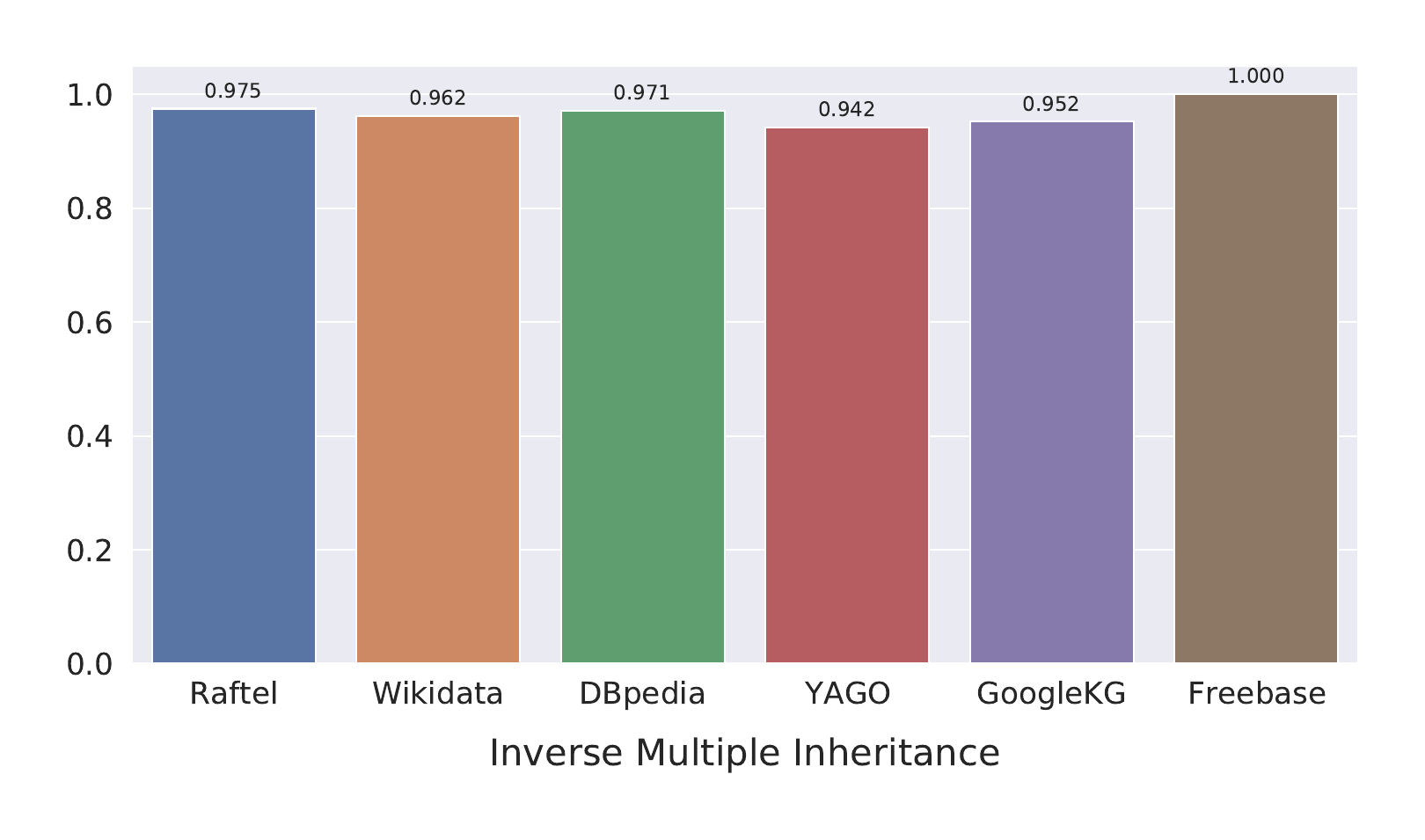}
\captionof{figure}{Inverse Multiple Inheritance (target language: Korean)}
\end{minipage}
\subsubsection{Subclass Property Acquisition}
\begin{minipage}{\linewidth}
\includegraphics[width=\linewidth]{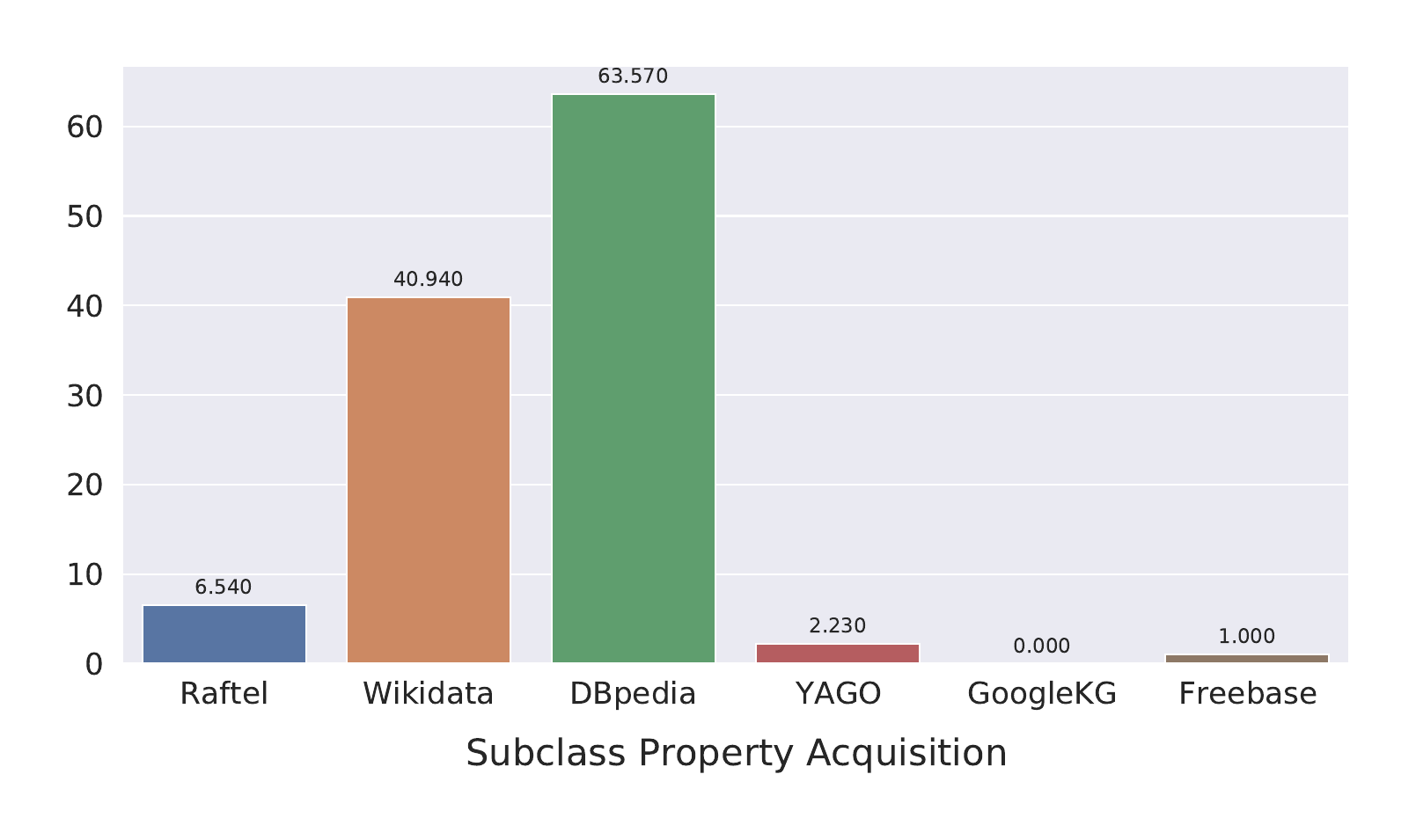}
\captionof{figure}{Subclass Property Acquisition (target language: Korean)}
\end{minipage}

\vspace{30mm}
\subsubsection{Subclass Property Instantiation}
\begin{minipage}{\linewidth}
\includegraphics[width=\linewidth]{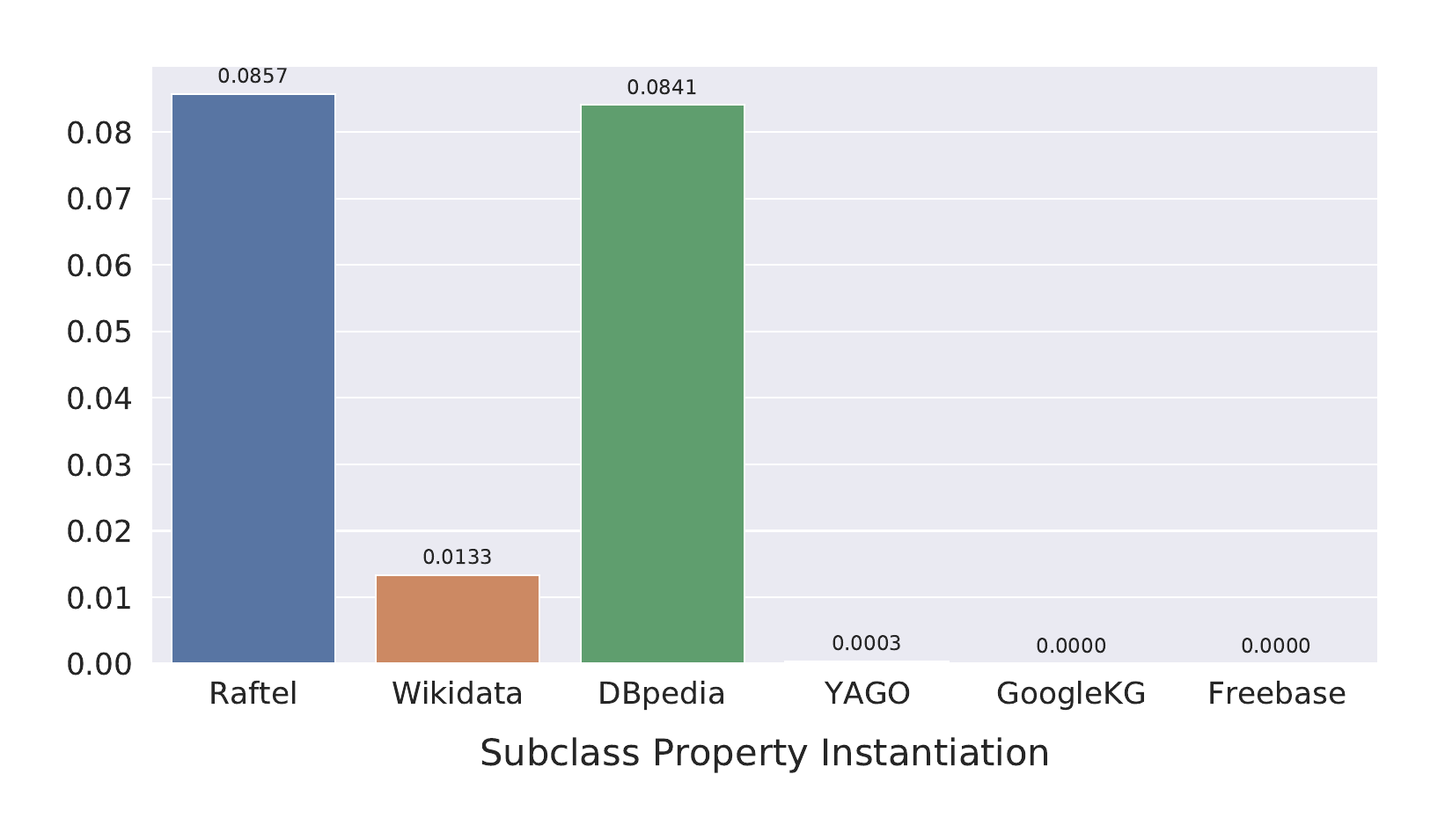}
\captionof{figure}{Subclass Property Instantiation (target language: Korean)}
\end{minipage}

\end{document}